\documentclass[journal, 10 pt, letterpaper, final]{IEEEtran} 

\IEEEoverridecommandlockouts                              

\usepackage{graphicx}
\usepackage{xcolor}

\newtheorem{definition}{Definition}
\newtheorem{theorem}{Theorem}
\newtheorem{corollary}{Corollary}[theorem]

\let\labelindent\relax 
\usepackage{enumitem}  

\usepackage{caption}
\usepackage{subcaption}
\usepackage{cite}

\newcommand{\calE}{{\cal E}}

\newcommand{\calG}{{\cal G}}

\newcommand{\calN}{{\cal N}}
\newcommand{\calO}{{\cal O}}

\newcommand{\calS}{{\cal S}}

\newcommand{\calV}{{\cal V}}

\newcommand{\calX}{{\cal X}}

\newcommand{\bfc}{\mathbf{c}}

\newcommand{\bfg}{\mathbf{g}}

\newcommand{\bfp}{\mathbf{p}}

\newcommand{\prl}[1]{\left(#1\right)}
\newcommand{\brl}[1]{\left[#1\right]}
\newcommand{\crl}[1]{\left\lbrace#1\right\rbrace}

\usepackage{amsmath} 
\usepackage{amsfonts}
\usepackage{amssymb}  
\usepackage{hyperref}

\title{
Resilient Consensus via \\Voronoi Communication Graphs
}

\author{Kelsey Saulnier$^{1}$, Lifeng Zhou$^{2}$, George Pappas$^{1}$, Vijay Kumar$^{1}$
\thanks{We gratefully acknowledge the support of ARL DCIST CRA W911NF-17-2-0181, NSF Grants CCR-2112665, and ONR grant N00014-20-S-B001.}
\thanks{$^{1}$The authors are with the GRASP Laboratory at the University of Pennsylvania, Philadelphia, PA, 19104, USA
	(email: {\tt\small saulnier, pappasg, kumar@seas.upenn.edu}).}
\thanks{$^{2}$The author is with the Department of Electrical and Computer Engineering, Drexel University, Philadelphia, PA 19104, USA (email: {\tt\small lz457@drexel.edu}).}
	}

\begin{document}

\maketitle
\thispagestyle{empty}
\pagestyle{empty}

\begin{abstract}
Consensus algorithms form the foundation for many distributed algorithms by enabling multiple robots to converge to consistent estimates of global variables using only local communication. However, standard consensus protocols can be easily led astray by non-cooperative team members. As such, the study of resilient forms of consensus is necessary for designing resilient distributed algorithms. W-MSR consensus is one such resilient consensus algorithm that allows for resilient consensus with only local knowledge of the communication graph and no a priori model for the data being shared. However, the verification that a given communication graph meets the strict graph connectivity requirement makes W-MSR difficult to use in practice. In this paper, we show that a commonly used communication graph structure in robotics literature, the communication graph built based on the Voronoi tessellation, automatically results in a sufficiently connected graph to reject a single non-cooperative team member. Further, we show how this graph can be enhanced to reject two non-cooperative team members and provide a roadmap for modifications for further resilience. This contribution will allow for the easy application of resilient consensus to algorithms that already rely on Voronoi-based communication such as distributed coverage and exploration algorithms.

\end{abstract}



\section{Introduction}
As multi-robot systems become larger, more complex, and operate over larger areas, there is a greater need for computation and coordination solutions that are distributed instead of centralized. One challenging objective in distributed systems is to ensure that the system is in agreement with the solution that is being computed so it is no surprise that consensus algorithms are often heavily featured in distributed computing and coordination. Consensus algorithms are used to allow agents to arrive at an agreement on estimates of variables in a distributed fashion and they appear in wide-ranging applications from distributed filtering~\cite{Olfati-Saber2007_kalman}, distributed field estimation~\cite{Schwager2015,Jang2020_distributedGP}, data aggregation~\cite{Kumar2004_dataaggregation}, formation control\cite{Ren2008}, flocking~\cite{Turgut2008_flocking}, and more~\cite{Ren2007_consensus_review}.

Consensus algorithms are sometimes used in cases where a leader is designated to lead the team to a desired value, such as in formation control or guided flocking where specific robots have special information which is used to guide the group~\cite{Su2009_flockingleader}. However, this sensitivity to leadership is problematic in the case where no leader is desired since a single malfunctioning, non-cooperative, or malicious agent can guide the entire network of agents' behavior. Several types of methods have been introduced to guard against such agents such as reputation or trust-based methods \cite{Liu2014}, where each node gains a score based on its behavior, and fault  detection and isolation methods such as \cite{Pasqualetti2007,Pasqualetti2012}. Most methods require each node to have significant knowledge of the communication graph structure. 
LeBlanc \emph{et al.} introduced the Weighted-Mean Subsequence Reduced (W-MSR) algorithm to address the problem of non-cooperative team members disrupting the consensus value, without the need for global information about the communication network or computationally complex algorithms~\cite{LeBlanc2013}. This powerful tool is limited by a strict network connectivity requirement called $(r,s)$-robustness, required to guarantee convergence of the consensus algorithm~\cite{LeBlanc2012} to a safe value. However,  verifying that a given network is sufficiently robust has been shown to be NP-Complete~\cite{Zhang2012_complexity_of_rrobust}, which makes this algorithm generally unsuitable for use in mobile robot teams where the communication network is continuously changing, or even for large static networks if computation time is limited. Prior work shows some exceptions such as~\cite{Saulnier2017} where an easily computed bound is used to allow mobile robots to use W-MSR consensus to agree on a flocking direction in the presence of malicious team members. However, this easily computed bound often results in highly connected communication graphs which can be unsuitable for applications where robots must be spread around the environment. Some specialized formation rules have also been studied, such as formations on lattices \cite{Saldana2018_triangle,Yu2020,Guerrero-Bonilla2020_lattice} and periodic formations \cite{Saldana2017}, with the former being limited to a single non-cooperative agent, and the later being limited to circular periodic patterns.

In this work, we expand the knowledge of graphs that can be used with the W-MSR algorithm. In particular, we will show that the network created by connecting the neighbors in a Voronoi tessellation is another special network that can be used with W-MSR consensus to reject a single non-cooperative member, and we will further provide a method for enhancing that communication graph to allow for two or more non-cooperative team members. The Voronoi tessellation is frequently used to divide work in multi-robot coverage and exploration tasks~\cite{Schwager2015, Breitenmoser2010_voronoicovex,Cortes2004,Luo2018_voronoiadaptive,Kemna2017_dynamic_voronoi,Hu2020_voronoiexplore}, so we believe this network and its resilience properties will be very useful for the application of resilient consensus to large mobile robot teams, by eliminating the need to explicitly compute the robustness of the communication graph.
\section{Background}
\label{sec:resilient_voronoi_background}
Let there be a set of robots $\calV$, such that $|\calV|=N$, and an undirected communication graph $\calG(\calV,\calE)$ such that if $(v_i,v_j)\in\calE$ the robots at $v_i$ and $v_j$ are \emph{neighbors} indicating that they communicate with each other, denoted $v_i\in\calN_j$ and $v_j\in\calN_i$. Weighted linear consensus is defined as
\begin{equation}
    \label{eq:linear_consensus}
    c^{(i)}[k+1] = w_{ii}c^{(i)}[k] + \sum_{j\in\calN_{i}} w_{ij}c^{(j)}[k],
\end{equation}
where $\sum_j w_{ij}=1$
which, when performed on a connected, undirected, communication graph $\calG$ results in the robots achieving asymptotic consensus on a weighted average of the initial values
\begin{equation*}
    \lim_{k\rightarrow\infty}c^{(i)}[k] = \sum_{i=1}^n \alpha_i c^{(i)}[0],
\end{equation*}
for weights $\alpha_i$ such that $\sum_i \alpha_i = 1$ at an exponential rate determined by the algebraic connectivity of the graph~\cite{Olfati-Saber2007_linear_consensus_convergence}. 

A feature of standard linear consensus is that a single robot can lead the network to reach a consensus value arbitrarily far from the the initial conditions by simply not cooperating with the consensus update~\cite{Gupta2006_convergence_follows}. If such a robot exists in the network, the remaining robots will converge to the value that the non-cooperative robot is sharing with its neighbors. This can cause arbitrarily bad behavior in the system if the controlling robot is malicious or if the non-cooperative behavior is not part of the design. 

The W-MSR algorithm is a modified version of the linear consensus which is designed to reject the influence of a number of non-cooperative robots in the network with only local information about the communication graph and with no model of the data the robots are reaching the consensus on. It is defined for a parameter $F$ as
\begin{equation}
    \label{eq:wmsr}
    c^{(i)}[k+1] = w_{ii}c^{(i)}[k] + \sum_{j\in\calN_{i,-F}[k]} w_{ij}c^{(j)}[k]
\end{equation}
where the set $\calN_{i,-F}[k]\subset\calN_i$ is constructed by ordering the set of consensus values provided by neighbors of robot $i$ and removing the neighbors corresponding to the $F$ largest values larger than $c^{(i)}[k]$, and the $F$ smallest values smaller than $c^{(i)}[k]$. This results in at most $2F$ neighbors being removed from the set at each round of W-MSR. Note that, for example, only $F$ neighbors are removed from the set if $\forall\;\crl{c^{(j)}:j\in\calN_i}$, $c^{(j)}[k]<c^{(i)}[k]$, since in that case all $F$ largest values are smaller than $c^{(i)}[k]$ and so they are not removed. W-MSR with parameter $F=0$ is simply the standard weighted linear consensus defined in \eqref{eq:linear_consensus}.

W-MSR consensus will allow robots in a sufficiently robust communication graph $\calG$ to reach consensus in the presence of \emph{non-cooperative} robots. Non-cooperative robots are any robots in the consensus network which are sharing values with their neighbors which were not computed using the consensus algorithm. We assume that non-cooperative robots share the same value with all their neighbors at each consensus step, but are otherwise unrestricted in their behavior. If the network is sufficiently robust, the cooperative robots are guaranteed to reach a \emph{safe} consensus value, defined as a value in between the maximum and minimum initial values of the cooperative robots,
\begin{equation}
\label{eq:safeconvergence}
    \lim_{k\rightarrow \infty} c^{(i)}[k] = c_s,\;\; \;\;\;\; \min{\bfc[0]}\leq c_s \leq \max{\bfc[0]}.
\end{equation}

The convergence of W-MSR is contingent on each robot in the communication graph $\calG$ having a sufficient quantity and diversity of communication, quantified by a graph property called $(r,s)$-robustness. 

\begin{definition}[$(r,s)-$robust]
\label{defn:rsrobust}
A communication network graph $\mathcal{G}(\calE,\calV)$ is $(r,s)$-robust if and only if for every pair of non-empty disjoint subsets $\calS_1, \calS_2\subset \calV$ at least one of the following holds
\begin{enumerate}
    \item $|\calX_{\calS_1}^r| = |\calS_1|$
    \item $|\calX_{\calS_2}^r| = |\calS_2|$
    \item $|\calX_{\calS_1}^r|+|\calX_{\calS_2}^r| \geq s$
\end{enumerate}
where $\calX_{\calS_k}^r = \left\lbrace v_i \in \calS_k \; : \; |\calN_i \setminus \calS_k| \geq r \right\rbrace$ is the number of nodes in $\calS_k$ with at least $r$ neighbors outside $\calS_k$.
\end{definition} 

The required level of robustness to guarantee convergence under W-MSR is determined by the threat model. If the total number of non-cooperative robots in the network $N_{nc}$ is bounded by parameter $F$, this is called the $F$\emph{-global} threat model. To guarantee convergence in this case the communication graph must be at least $(F+1,F+1)$-robust. Alternatively under the $F$\emph{-local} threat model there are no more than $F$ non-cooperative robots in the neighbor set of every cooperative robot. This can result in a $N_{nc}>>F$. In the $F$-local case the graph must be at least $(2F+1,1)$-robust. The non-cooperative robots can fail to reach consensus or be corrupted by the non-cooperative robots if the graph does not have a sufficiently high level of $(r,s)$-robustness. This failure to converge can happen when using W-MSR with parameter $F>0$ even in the case where $N_{nc}=0$ if the graph is not sufficiently robust, since robots can have so few neighbors that $\calN_{i,-F}=\emptyset$ at all times. 

The $(r,s)$-robustness is coNP-Complete to verify for a given $r$, $s$, and communication network $\calG$~\cite{Zhang2012_complexity_of_rrobust} and therefore it is useful to consider cases where sufficient robustness can be guaranteed without requiring explicit verification. This is particularly important in mobile robot systems, where communication graphs may vary over time, and in systems with large numbers of robots to deploy where even one time verification can take significant computation time.

In this paper, we add to the state of the art by showing that the graph created by connecting neighbors in a Voronoi tessellation, also called the Delaunay triangulation, is $(2,2)$-robust and therefore can support W-MSR with $F=1$, to reject a single non-cooperative robot under the $F$-global threat model. Furthermore, we prove that the Voronoi graph can be augmented by connecting two hop neighbors to achieve $(3,3)$-robustness, required for handling an $F$-global threat with $F=2$, or $F$-local threat with $F=1$. Finally, we show numerical evidence that this augmentation scheme can be continued to create networks with increasing robustness guarantees.

\section{Resilient Consensus for Voronoi Graphs}
\label{sec:resilient_voronoi_theorems}
\begin{figure}
    \centering
    \includegraphics[width=.35\textwidth]{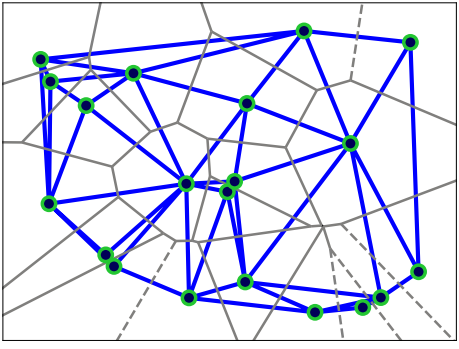}
    \caption{\small Robot positions are shown in green, the current Voronoi partition in grey, and the communication graph $\calG$ calculated using Delaunay triangulation is shown in blue. }
    \label{fig:voronoi_delaunay_diagram}
\end{figure}

In this section, we prove that the communication graph created by connecting each robot to the robots in neighboring Voronoi cells creates a graph, $\calG_\Delta$, that is resilient to a single non-cooperative robot under the F-global threat model. An example of this graph is shown in Figure~\ref{fig:voronoi_delaunay_diagram}. This result is related to previous work by Salda\~{n}a \emph{et al.} where the authors argue that all triangular graphs are $(2,2)$-robust~\cite{Saldana2018_triangle}. There are technical challenges that prevent us from directly using their result, so we provide a proof specific to Voronoi communication graphs. Further, we prove that this graph can be enhanced by connecting two-hop neighbors to create a graph that is resilient to 2 non-cooperative robots under the F-global threat model or 1 non-cooperative robot under the F-local threat model. Lastly, we show numerical evidence that this graph enhancement method can be used to produce graphs that are resilient against any given number of non-cooperative robots, assuming the graph contains sufficient vertices.

\begin{theorem}
\label{thm:voronoi_robust}
The communication graph, $\calG_\Delta(\calV,\calE)$, formed by the Delaunay triangulation with $N>2$ robots is $(2,2)-$robust.
\end{theorem}

In the following proof, and other proofs in this paper we will use the notation $\calG[\calS]$ to indicate the \emph{induced subgraph} of $\calG(\calV,\calE)$ formed by the vertex set $\calS$. This is the subgraph containing vertices $\calS\subset\calV$ and all edges in $\calE$ which have both endpoints in $\calS$. We use $\prl{\bullet}$ to indicate an ordered path of vertices and $\crl{\bullet}$ to indicate an unordered set of vertices. 

\begin{IEEEproof}
 Given any pair of non-empty disjoint subsets $\calS_1,\calS_2\subset\calV$, let $\crl{\calS_{a,i}}$ be the set of connected components of the induced subgraph $\calG_\Delta[S_a]$, where $\calS_a\in\crl{\calS_1,\calS_2}$. There are then the following cases:
\begin{enumerate}
    \item $\exists \calS_a$ such that $|\calS_{a,l}| = 1,\; \forall l$. All connected components of $\calG_\Delta[\calS_a]$ are size 1. Since all nodes have degree $\geq 2$ (triangle faces), this means that each subset of size 1 has at least 2 edges leaving the set and therefore $|\calX_{\calS_a}^2| = |\calS_a|$, satisfying Condition 1 or 2 in Definition \ref{defn:rsrobust}.
    \item $\exists l_1,l_2, \; s.t., \;|\calS_{1,l_1}| > 1$ and $|\calS_{2,l_2}| > 1$, i.e. $\calG_\Delta[\calS_1]$ and $\calG_\Delta[\calS_1]$ each contain a connected component with multiple vertices. First consider $\calS_{2,l_2}$. We know there exists a path $\prl{ v_i, v_j, ... , v_g }$ from $\calS_1$ to $\calS_{2,l_2}$ such that $v_i\in \calS_1$, $v_j\in \bar{\calS_1}$, the compliment of $\calS_1$, and $v_g\in\calS_{2,l_2}$. A diagram of this path can be seen at the top of Figure \ref{fig:proof22_pathandlogic}. We know $v_j$ has a neighbor $v_l\in\bar{\calS_1}$ (since $\calS_{2,l_2}$ is a connected component of at least two vertices, worst case $v_l$ can be the other node in this connected component). The bottom panel of Figure \ref{fig:proof22_pathandlogic} shows this path. Since $\calG$ is a triangulation, all the interior faces must be triangles. Thus there exists an ordering of the neighbors of $v_j$ such that $\crl{T_{j,n,m}}, \; n={1,...,|\calN_{v_j}|-1},\; m={2,...,|\calN_{v_j}|}$ are triangles (notation $T_{j,n,m}$ is a triangle face with vertices $v_j,v_n,v_m$. Note if $v_j$ is not on the convex hull, there is another triangle $T_{j,|\calN_{v_j}|,1}$, but it is not needed in the proof). This implies that the neighbors of $v_j$ form a connected subset which is a simple path or cycle containing all neighbors. Thus, since $v_l$ and $v_i$ are both neighbors of $v_j$, there exists a path $\prl{v_l,...,v_n,v_m...,v_i}\in\calN_{v_j}$ such that $v_n\in\bar{\calS_1}$, $v_m\in\calS_1$. Therefore $v_m\in \calX_{\calS_1}^2$ with neighbors $\crl{v_j,v_n} \in\bar{\calS_1}$, proving that $|\calX_{\calS_1}^2|\geq 1$. Following the same logic with $\calS_{1,l_1}$ produces the result $|\calX_{\calS_2}^2|\geq 1$, therefore $|\calX_{\calS_1}^2|+|\calX_{\calS_2}^2|\geq 2$, satisfying Condition 3 in Definition \ref{defn:rsrobust} for $r=s=2$.
\end{enumerate}
\begin{figure}
    \centering
    \includegraphics[width=0.25\textwidth]{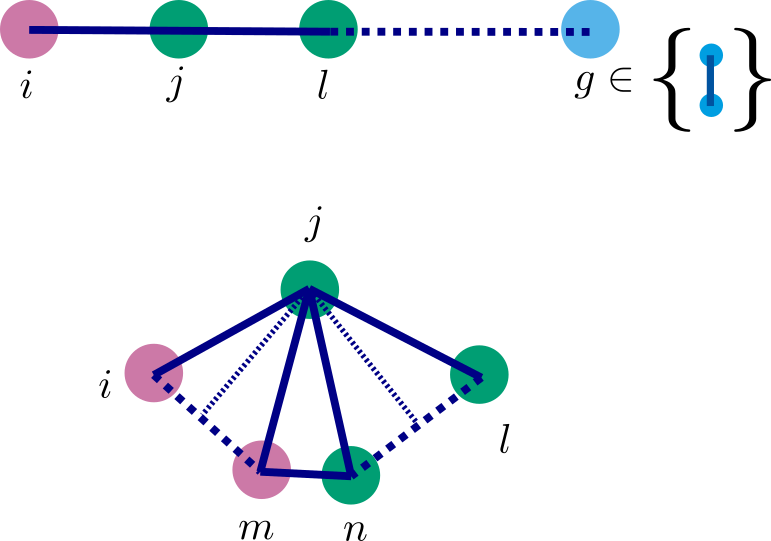}
    \caption{\small When $\calG[\calS_1]$ and $\calG[\calS_2]$ each have at least one connected component with 2 or more vertices, there exists a path from a vertex $v_i\in\calS_1$ (pink) to a vertex $v_g\in\calS_2$ (blue) such that $|\calN_g\cap\calS_2|\geq2$. The first vertex along this path in $\bar{\calS_1}$ (green), $v_j$, has a neighbor in $\calS_1$ and one in $\bar{\calS_1}$. Since each vertex's neighbors form a path (denoted by dotted lines), we know there must be a transition vertex $v_m\in\calS_1\in\calX_{\calS_1}^2$, i.e., having 2 neighbors outside $\calS_1$. }
    \label{fig:proof22_pathandlogic}
\end{figure}
Since for ever pair of disjoint subsets of $\calV$ one of the two cases holds, the graph satisfies the conditions in Definition~\ref{defn:rsrobust} with $r=2,s=2$ for all pairs of disjoint subsets and is $(2,2)$-robust.

 \end{IEEEproof}

\begin{figure}
    \centering
    \begin{subfigure}[b]{0.23\textwidth}
    \includegraphics[width=\textwidth]{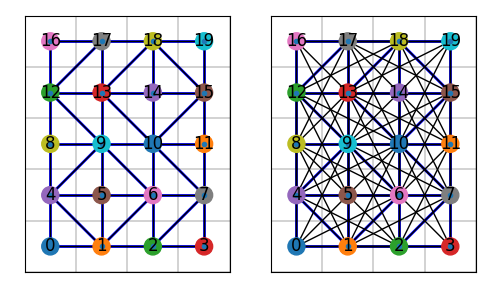}
    \caption{grid}
    \label{fig:special_voronoi_formations_grid}
    \end{subfigure}
    \begin{subfigure}[b]{0.23\textwidth}
    \includegraphics[width=\textwidth]{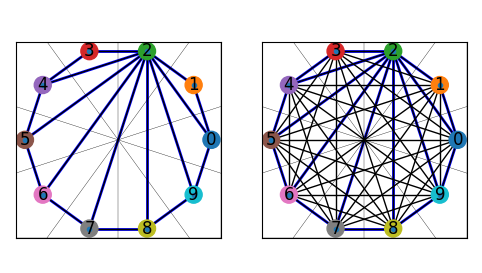}
    \caption{circle}
    \label{fig:special_voronoi_formations_circle}
    \end{subfigure}
    \begin{subfigure}[b]{0.23\textwidth}
    \includegraphics[width=\textwidth]{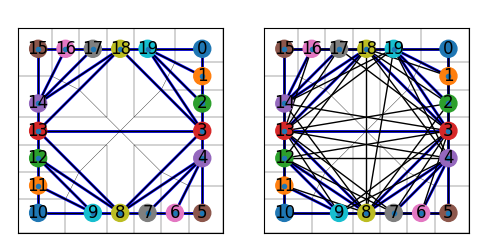}
    \caption{square}
    \label{fig:special_voronoi_formations_square}
    \end{subfigure}
    \begin{subfigure}[b]{0.23\textwidth}
    \includegraphics[width=\textwidth]{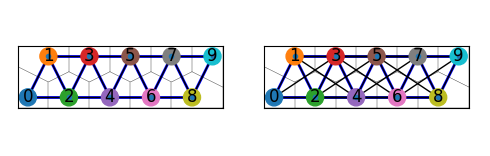}
    \caption{two lines}
    \label{fig:special_voronoi_formations_twolines}
    \end{subfigure}
    \caption{\small A comparison of $\calG_\Delta$ (left figure in each panel) and $\calG_{\Delta 2}$ (right figure) on for some interesting robot formations. The lines of the Voronoi tessellation are shown in light grey. (a) shows the robots on a grid which results in local connections between robots, with neighbors being only one row away for $\calG_\Delta$, and up to two rows away for $\calG_{\Delta 2}$. In contrast, (b) shows a circular formation which results in a fully connected graph for $\calG_{\Delta 2}$. In (c), a hollow square formation is shown. Finally, in (d), a formation of two lines is shown which contains very few edges to nearby robots which, like (a) increase incrementally from $\calG_\Delta$ to $\calG_{\Delta 2}$.}
    \label{fig:special_voronoi_formations}
\end{figure}

The graph $G_\Delta$ can be used with W-MSR with parameter $F=1$ to reject a single $F$-global non-cooperative robot. We will now provide a method to increase the $(r,s)$-robustness and reject a larger set of non-cooperative robots. To this end we define an enhanced graph, $\calG_{\Delta 2}$, which is the graph $\calG_\Delta$ with additional edges to directly connect 2-hop neighbors. A comparison of $\calG_\Delta$ and $\calG_{\Delta 2}$ for several interesting formations can be seen in Figure \ref{fig:special_voronoi_formations}.

\begin{theorem}
\label{thm:voronoi_robust_33}
If the communication graph $\calG_\Delta(\calE_\Delta,\calV_\Delta)$ formed by connecting the Voronoi neighbors with $N>4$ robots is extended to $\calG_{\Delta 2}(\calE_{\Delta 2},\calV_\Delta)$ with $\calE_\Delta \subset \calE_{\Delta 2}$ such that if $(v,w)\in\calE_\Delta$ and $(w,z)\in\calE_\Delta$ then $(v,z)\in\calE_{\Delta 2}$ (connecting nodes in $\calG_\Delta$ to their neighbor's neighbors), then the resulting graph is $(3,3)-$robust.
\end{theorem}

The proof of this theorem is included in the Appendix.

\begin{corollary}
The graph $\calG_{\Delta 2}$ when $F=1$ is also resilient to $F$-\emph{local} non-cooperative robots, where $F$-local indicates fewer than $F$ non-cooperative neighbors in the neighborhood of \emph{every} robot.
\end{corollary}
\begin{IEEEproof}
LeBlanc et al.~\cite{LeBlanc2012} show that $(r,s)$-robustness is a generalization of $r$-robustness, such that a $(r,1)$-robust graph is $r$-robust. Since $\calG_{\Delta 2}$ is $(3,3)$-robust, that means it is $3$-robust. A $2F+1$-robust graph is resilient to one F-local non-cooperative robot. Thus $\calG_{\Delta 2}$ is resilient to $F=1$ F-local threats.
\end{IEEEproof}

Considering that we have shown that the Voronoi graph results in $(2,2)$-robustness and the connection of 2-hop neighbors results in $(3,3)$-robustness, it is interesting to consider whether this method of increasing robustness would continue. For example, if one were to create a graph $\calG_{\Delta 3}$ from $\calG_{\Delta 2}$ by adding edges between the 3-hop neighbors of $\calG_\Delta$ such that $\calE_\Delta\subseteq\calE_{\Delta2}\subseteq\calE_{\Delta 3}$, would the resulting graph be $(4,4)$-robust, and so on. The corresponding proofs would get rapidly more complex, however in Section~\ref{sec:enhanced}, we examine this relationship in simulation with small graphs where the $(r,s)$-robustness can be computed without a large computation time burden.

\section{Examples}
\label{sec:resilient_voronoi_examples}
Here we will provide three examples of how this result could be used. For all examples we specify the parameter $F$ used for W-MSR, with plain consensus being $F=0$, and use \eqref{eq:wmsr} with equal weights $w_{i,j}=1/|\calN_{i,-F}+1|$  $\forall j\in\calN{i,-F}\cap\crl{i}$. The first example is distributed parameter estimation, where the robots are attempting to agree on a parameter while they are distributed over a large area, for example in a coverage task. This example shows the basic function of resilient consensus in a scenario where the Voronoi tessellation might naturally be used. The second example is convergence to a rendezvous location, where the robots run consensus to agree on rendezvous coordinates and travel to the agreed-upon location. This example shows how the Voronoi graph can be used to create resilient communication graphs even in scenarios where Voronoi tessellations are not typically used. Lastly, we show a distributed mapping example where resilient consensus enables the robots to successfully map an environment while a non-cooperative team member attempts to block the exploration of part of the environment. We provide a video of the examples with moving robots to supplement the figures in this section.\footnote{\url{https://youtu.be/zZ5RLdVfUEY}}

\subsection{Distributed Parameter Estimation}
In this example, a set of $N$ robots are distributed over an area and are tasked with reaching consensus on a parameter $\alpha$ for which each robot has an estimate $\alpha_i$. In the network there are some number, $N_{nc}$, of non-cooperative robots. In this simple example, the robots are stationary during consensus which will allow for an investigation of the graphs $\calG_\Delta$ and $\calG_{\Delta 2}$.

\begin{figure}
    \centering
    \includegraphics[width=0.5\textwidth]{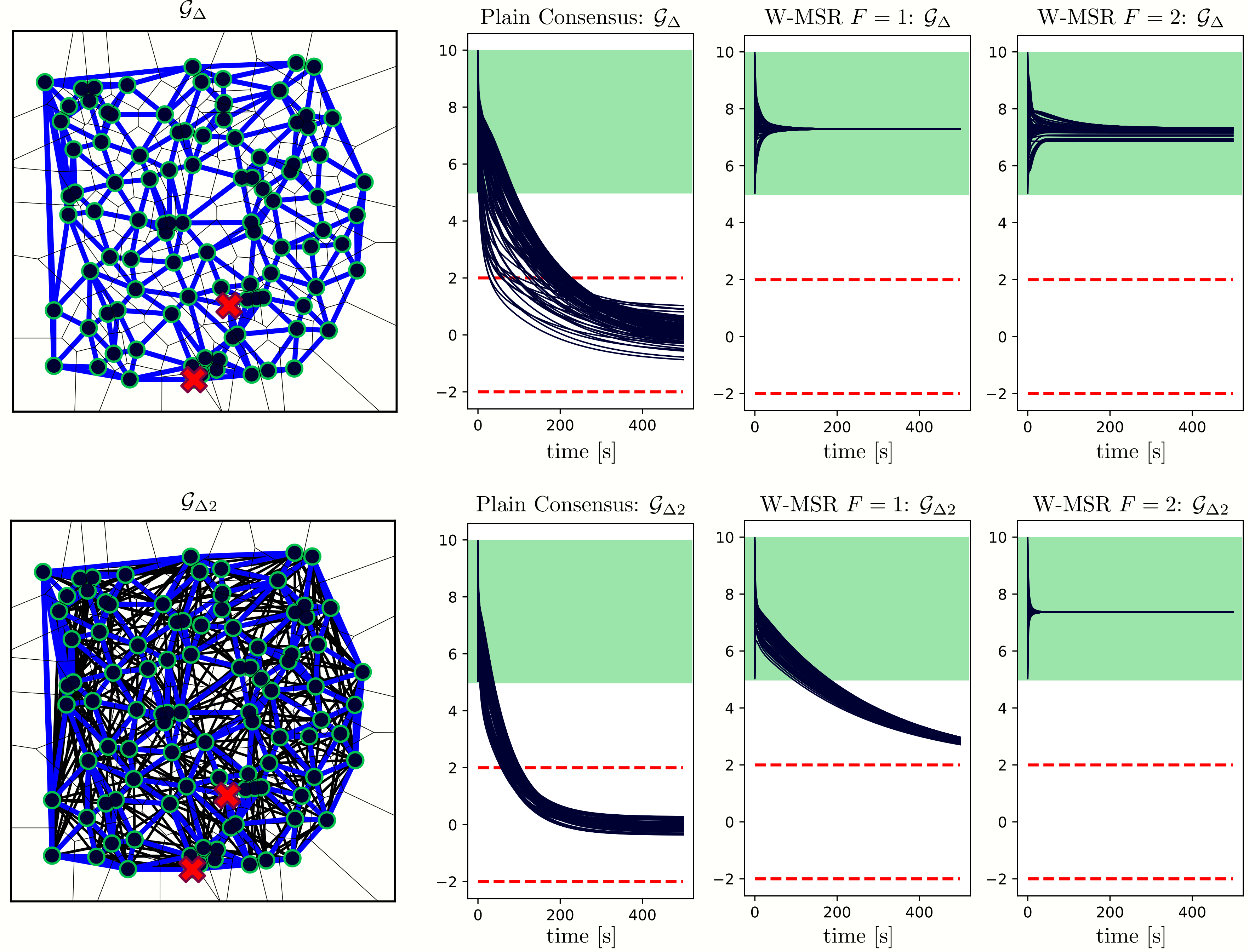}
    \caption{\small Consensus in a network of $N=100$ agents, $N_{nc}=2$ of which are non-cooperative and indicated by red `X's. Non-cooperative agents share constant values with their neighbors, indicated by red dashed values in the consensus plots. The green area of the plots indicates the range of \emph{safe} consensus values as defined in \eqref{eq:safeconvergence}}
    \label{fig:consensus_many}
\end{figure}

 Figure \ref{fig:consensus_many} shows a group of $N=100$ robots randomly distributed in a coverage area. In this case, there are $N_{nc}=2$ non-cooperative robots which share values outside the safe convergence region to their neighbors shown as the red dotted lines on the plots. It can be clearly seen that the plain consensus does not provide a safe value or consensus among the cooperative robots as they each receive different amounts of influence from the two non-cooperative robots. When using W-MSR with $F=1$, $F$ is too small to handle the $N_{nc}=2$ F-global threat. Despite this, the cooperative robots are able to reach a consensus on a safe value on $\calG_\Delta$. This is likely due to the fact that no cooperative robot has multiple non-cooperative neighbors. On $\calG_{\Delta 2}$, however, the cooperative agents are converging to the value of one of the non-cooperative robots. This shows an interesting interplay between connectivity and resilience. The greater connectivity of $\calG_{\Delta 2}$ provides greater resilience, but if $F<N_{nc}$, it can also increase the number of robots that have more than $F$ non-cooperative neighbors, causing W-MSR to fail. When using an appropriate parameter $F=N_{nc}=2$, $\calG_{\Delta 2}$ is able to reject both non-cooperate agents. This can be seen on the far right of Figure \ref{fig:consensus_many}. In contrast, $\calG_\Delta$ does not provide enough connectivity to successfully run W-MSR with $F=2$, which is evidenced by a single cooperative robot with a constant consensus value. That robot has only 2 neighbors which it will always discard with parameter $F=2$ and therefore it never deviates from its initial value.

\begin{figure}
    \centering
    \includegraphics[width=0.35\textwidth]{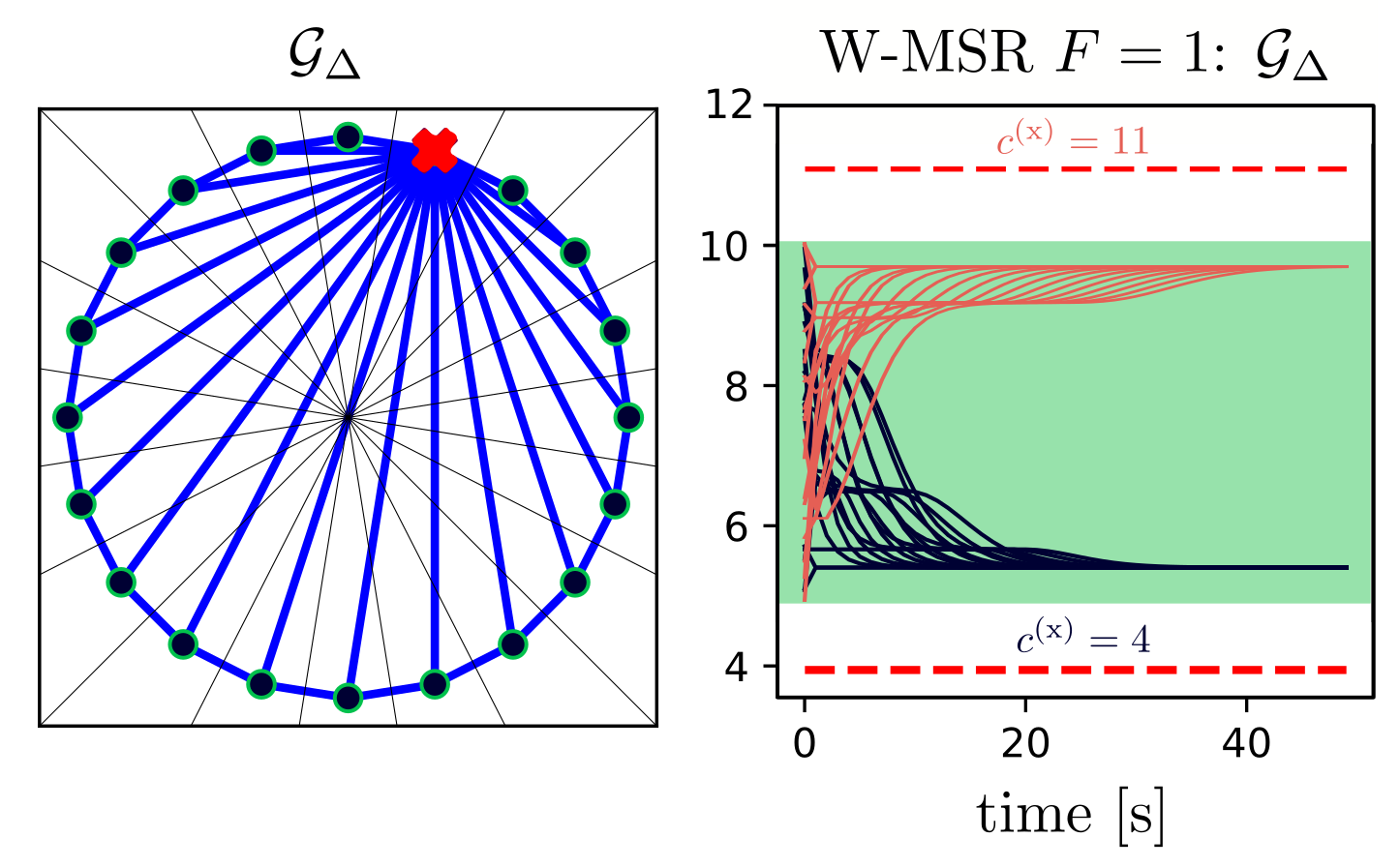}
    \caption{\small The non-cooperative agent can influence the final consensus value within the safe range defined by the initial cooperative values, indicated by the green area of the consensus graph. The consensus graph (right) shows the results of two consensus simulations on the graph shown on the left. In each case, the cooperative robots begin with the same initial values, but the non-cooperative robot, marked with a red `x', shares a different constant value $c^{(\text{x})}=11$ or $c^{(\text{x})}=4$. This results in very different consensus values though both are within the safe range defined by the cooperative initial conditions.}
    \label{fig:circle_above_below}
\end{figure}

We also examine the influence the non-cooperative robot is able to exert on the final consensus value of the non-cooperative robots when using resilient consensus. Figure \ref{fig:circle_above_below} shows a network of $N=20$ agents arranged in a circular configuration with $N_{nc}=1$. This example shows the effect the non-cooperative agent can have on the cooperative agents in a worst-case scenario. In this case, using $\calG_\Delta$, the non-cooperative agent is neighbors with all the other agents, and the cooperative agents each have only 2 cooperative neighbors. We know from the use of $\calG_\Delta$ that the network is $(2,2)$-robust, so the value the non-cooperative agent shares will be discarded under the W-MSR algorithm. However, the cooperative agents discard good values as well. In the case where the non-cooperative agent is broadcasting a value lower than all the other agents, each cooperative agent will discard that value, and if a neighboring agent shared a value \emph{larger} than the agent's current value it will discard that value too. In this case, this means that the cooperative values discarded by W-MSR are always larger, which influences the final consensus value lower, in this case to $5.4$. Likewise, a non-cooperative value larger than all the cooperative values will influence the final consensus value higher, in this case to $9.5$. The range of this influence is limited by the range of the initial cooperative values, whereas without resilient consensus the influence of the non-cooperative robot is unbounded.

\subsection{Polygon Rendezvous}
In this example, a set of $N$ robots are distributed in the environment and must come to an agreement on a rendezvous position. Each robot has an assigned offset and angle from the agreed rendezvous point, with the locations corresponding to the corners of an $N$-sided regular polygon. At each time step, the robots run consensus using their current Voronoi tessellation to determine their neighbors according to $\calG_\Delta$ or $\calG_{\Delta 2}$. Robots seek to find consensus on the rendezvous center location coordinates $(c_x,c_y)$ individually:
\begin{equation*}
    c_x^{(i)}[k+1] = \frac{1}{|\calN_{i,-F}|+1}\prl{ c_x^{(i)}[k] + \sum_{j\in\calN_{i,-F}} c_x^{(j)}[k]}
\end{equation*}
and the same for $c_y^{(i)}$, where $\calN_{i,-F}$ is the set of neighbors after discarding for W-MSR (with the plain consensus being $F=0$ indicating no reduction of the neighbor set). The robots then compute their current goal location from their estimate of the rendezvous center as
\begin{equation*}
    \brl{\begin{matrix} g_x^{(i)}[k]\\g_y^{(i)}[k] \end{matrix}} = \brl{\begin{matrix} c_x^{(i)}[k] + r\cos\prl{\frac{2\pi}{N}i}\\c_y^{(i)}[k] + r\sin\prl{\frac{2\pi}{N}i} \end{matrix}}
\end{equation*}
for a preassigned radius $r$ and robot number $i$. This distributes the robots to preassigned locations relative to the rendezvous point. Each robot then moves towards the rendezvous point with a basic proportional controller with a max velocity:
\begin{equation*}
    \bfp^{(i)}[k+1] = \bfp^{(i)}[k]+\tau\min\prl{\|\bfg^{(i)}[k]-\bfp^{(i)}[k]\|, v_{\max}}
\end{equation*}
for some time step $\tau$, robot positions $\bfp^{(i)} = \brl{p_x^{(i)},p_y^{(i)}}$, and goal positions $\bfg^{(i)} = \brl{g_x^{(i)},g_y^{(i)}}$.

\begin{figure*}[htbp]
    \centering
    \includegraphics[width=0.99\textwidth]{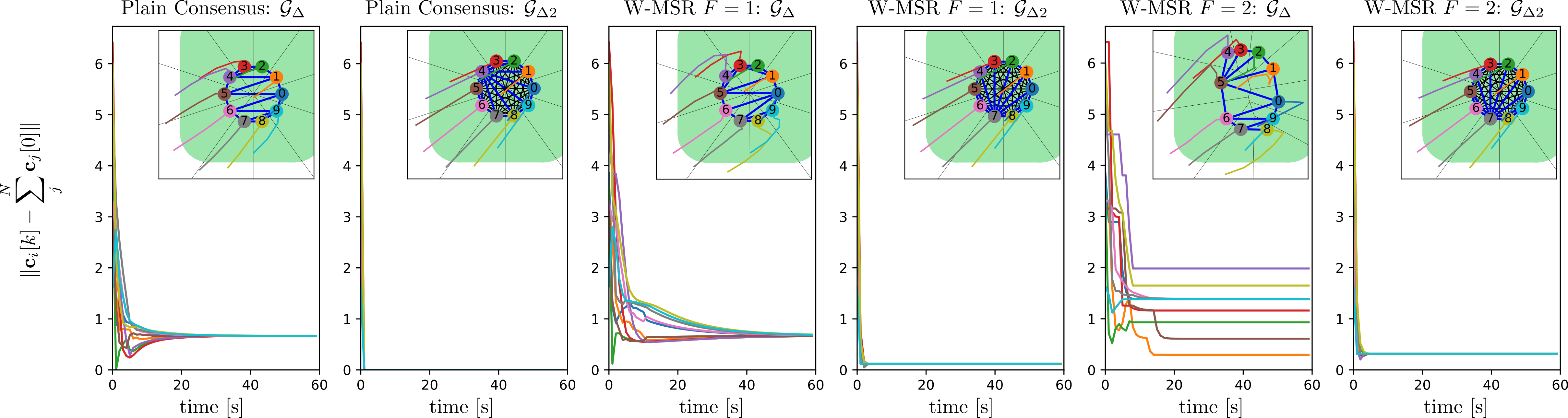}
    \caption{\small Consensus to a rendezvous location with only cooperative robots. The area marked in green shows the safe convergence region for the robots. The graph shows the difference from the average initial condition and is used to show convergence of the 2D polygon center. The robots reach consensus in all cases except when using W-MSR with $F=2$ on $\calG_{\Delta}$. In the failure case, the consensus is not reached because there is not enough connectivity in $\calG_\Delta$ to run W-MSR with $F=2$ since $\calG_\Delta$ allows robots to have as few as 2 neighbors.}
    \label{fig:polygon_nobad}
\end{figure*}

The first polygon rendezvous experiment explores the effects of W-MSR when there are no non-cooperative robots in the team. The results can be seen in Figure \ref{fig:polygon_nobad}. It shows that the consensus converges in all cases except the case where W-MSR with $F=2$ is being run on the Voronoi graph $\calG_\Delta$ which does not have enough connectivity to support $F=2$.


\begin{figure*}[htbp]
    \centering
    \includegraphics[width=0.99\textwidth]{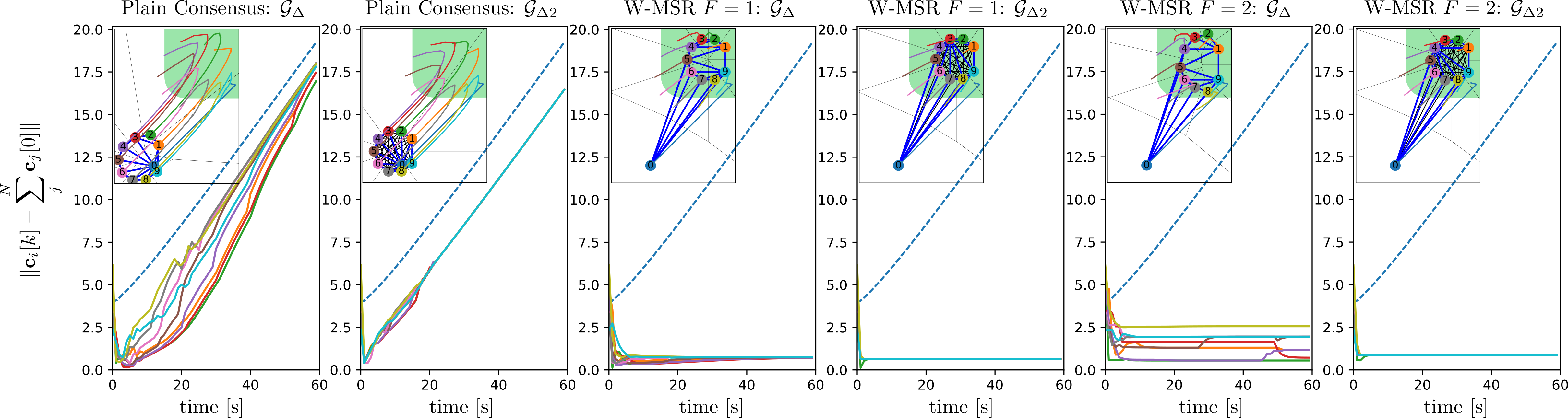}
    \caption{\small Consensus to a rendezvous location with $N_{nc}=1$. The non-cooperative robot (dotted line, blue 0 circle) begins sharing a random safe center location which it moves farther away from the safe region at each step. The safe region for the robots is shown in green. For plain consensus the robots follow the non-cooperative to the position it is communicating. For W-MSR, consensus is reached with $F=1$ and with $F=2$ on $\calG_{\Delta 2}$. In the case $F=2$ with communication graph $\calG_\Delta$, consensus is not reached because there is not enough connectivity. This results in a non-circular final formation.}
    \label{fig:polygon_1bad}
\end{figure*}

The second polygon rendezvous experiment introduces a single non-cooperative robot into the team which shares a constantly changing rendezvous location with its neighbors in an attempt to lead them away from the safe green area defined by the cooperative initial conditions. The results can be seen in Figure \ref{fig:polygon_1bad}. In this case, plain consensus follows the non-cooperative robot away from the safe convergence zone. This is the basic leader-follower behavior that consensus is occasionally used for, however, in this case, it is not the desired behavior because the leader is non-cooperative. For the cases where W-MSR is being run with $F=1$, the non-cooperative robot's rendezvous information is rejected and the cooperative robots are able to reach a consensus on a safe value. The same is true for W-MSR with $F=2$ on $\calG_{\Delta 2}$, though as before it cannot converge on $\calG_\Delta$ due to lack of connectivity.

\begin{figure*}[htbp]
    \centering
    \includegraphics[width=0.99\textwidth]{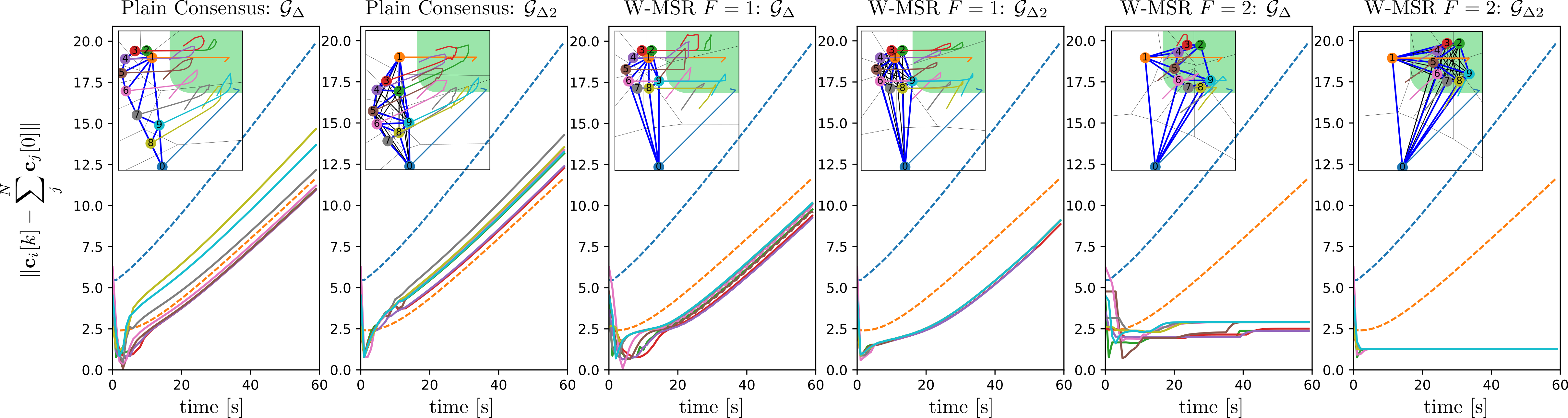}
    \caption{\small Consensus to a rendezvous location with $N_{nc}=2$. Non-cooperative robots (dotted lines, blue 0 and orange 1 circles) initially share a safe center point for consensus, then move their point farther from the safe area with each step. For plain consensus and for W-MSR with $F=1$, the robots follow the non-cooperative robots and since they are communicating different values, the cooperative robots do not reach consensus. In the W-MSR $F=1$ case, the robots successfully reject one of the non-cooperative robots but $F=1<N_{nc}$ is not sufficient to reject both. For W-MSR with $F=2$, the consensus is reached only on $\calG_{\Delta 2}$. In the case $F=2$ with communication graph $\calG_\Delta$, the consensus is not reached because there is not enough connectivity resulting in a non-circular final formation.}
    \label{fig:polygon_2bad}
\end{figure*}

Finally, the same experiment was run with 2 non-cooperative robots attempting to lead the team away from the safe consensus area, the results of which can be seen in Figure~\ref{fig:polygon_2bad}. In this case, the robots follow the non-cooperative robots in the plain consensus and the W-MSR consensus with $F=1$. In the plain consensus, the robots end up not reaching consensus on $\calG_\Delta$ though most cooperative robots closely follow the orange non-cooperative robot, which is the one with the most neighbors in $\calG_\Delta$. With $\calG_{\Delta 2}$, the robots have much higher connectivity and reach values closer to the average of the non-cooperative center locations. For W-MSR with $F=1$, the algorithm is able to reject one of the non-cooperative robots but not both, resulting in the cooperative robots following one of the non-cooperative robots away from the safe zone. Finally, when running W-MSR with $F=2$, the robots are able to reach a consensus on a safe value on the enhanced graph $\calG_{\Delta 2}$, but not on the lower connectivity triangulation graph $\calG_\Delta$.

\subsection{Map Consensus}
\begin{figure*}
\captionsetup[subfigure]{justification=centering}
    \centering
    \begin{subfigure}[t]{0.15\textwidth}
    \includegraphics[width=\textwidth]{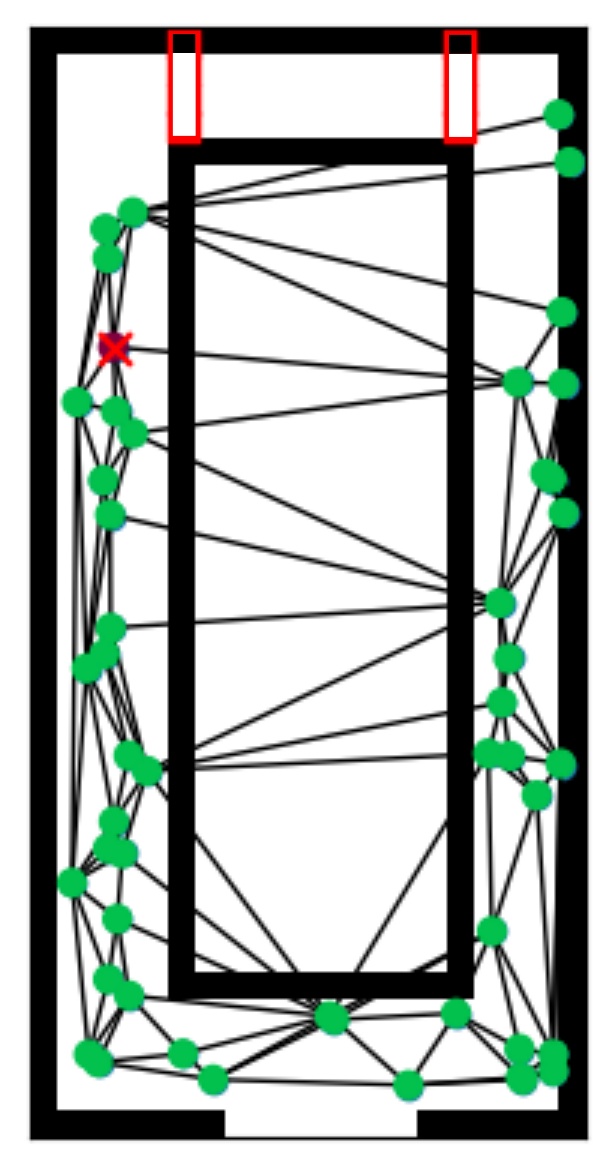}
    \caption{Environment}
    \label{fig:hallway_desc}
    \end{subfigure}\hspace{0.05\textwidth}
    \begin{subfigure}[t]{0.15\textwidth}
    \includegraphics[width=\textwidth]{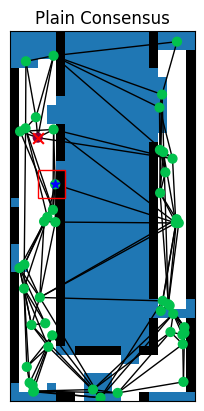}
    \caption{5 steps,\\ no W-MSR}
    \label{fig:hallway_5_0}
    \end{subfigure}
    \begin{subfigure}[t]{0.15\textwidth}
    \includegraphics[width=\textwidth]{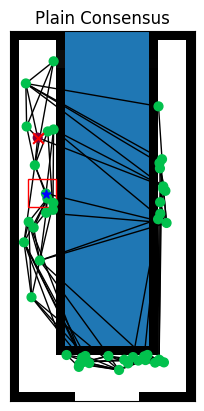}
    \caption{26 steps,\\ no W-MSR}
    \label{fig:hallway_26_0}
    \end{subfigure}\hspace{0.05\textwidth}
    \begin{subfigure}[t]{0.15\textwidth}
    \includegraphics[width=\textwidth]{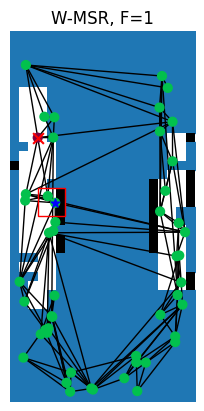}
    \caption{5 steps,\\ W-MSR F=1}
    \label{fig:hallway_5_1}
    \end{subfigure}
    \begin{subfigure}[t]{0.15\textwidth}
    \includegraphics[width=\textwidth]{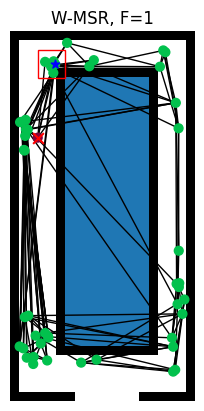}
    \caption{26 steps,\\ W-MSR F=1}
    \label{fig:hallway_26_1}
    \end{subfigure}
    \caption{\small Robots in a hallway environment move to explore unknown cells (blue) and communicate with their neighbors to reach a consensus on the occupancy map of a hallway environment. (a) shows the ground truth environment. The non-cooperative robot, indicated with a red `x', will claim the cells outlined in red are occupied. (b) and (c) show the occupancy map from the perspective of one of the cooperative robots (marked with a blue star) while convergence is in progress and after convergence, at 5 consensus steps and 26 consensus steps respectively. The red outline around the example robot indicates its sensor range. The robots running plain consensus believe that the hallway at the top is blocked. In (d) and (e) we see that using W-MSR with $F=1$ allows the cooperating robots to reject the false walls and fully explore the environment. }
    \label{fig:hallway}
\end{figure*}

The final example we will show is a case of map consensus. In this scenario, a team of robots is looking to achieve consensus on an occupancy map of a simple hallway environment. A non-cooperative robot is attempting to influence the team by sharing values indicating that part of the hallway is inaccessible. Each cell's occupancy is represented as a number $c_o^{(i)}\in\brl{0,1}$ where $0$ indicates free space, $1$ occupied space, and the numbers between are used to express uncertainty. In this example, the robots sense their environment noise-free in a square area around their location. The robots communicate according to the communication graph $\calG_\Delta$, and utilize a simple controller which moves them towards the nearest unknown cell that can be reached. If the robots perceive that there are no more reachable unknown cells, they consider the exploration complete and stop their motion. Unlike previous examples, robots in this example begin with no initial value for many of the cells in the environment. Robots keep track of which cells they do not know and run consensus only after they receive information about a cell from a neighbor. In the W-MSR case, this requires at least one neighbor to exist in the reduced neighbor set $\calN_{i,-F}$. Robots with no information use a neutral value of $0.5$ as their value when deciding which neighbors to disregard. Figure \ref{fig:hallway_desc} shows the hallway environment populated by cooperative robots (green) and one non-cooperative robot (red X). The non-cooperative robot is communicating with its neighbors that cells circled in red are occupied. Figures \ref{fig:hallway_5_0} and \ref{fig:hallway_5_1} show how the understanding of the environment has progressed after 5 consensus steps from the perspective of an example robot marked in the figures with a blue star. Two effects are clearly seen here. There are more cells in the W-MSR case which are still unknown to the example robot. These are indicated in blue. This is because the robot is unable to incorporate new information about a cell until there is sufficient information from its neighbors to run the W-MSR algorithm, i.e., W-MSR will provide no information if only one neighbor has information or if only two neighbors have information and it is conflicting. The robots running plain consensus exercise no such caution and it can be seen that the example robot believes there are walls where the non-cooperative agent is asserting there are walls. By 13 steps in the plain consensus case, and 26 steps in the W-MSR case, the example robot believes that it has a full understanding of the environment. It takes 36 steps for all robots in the W-MSR case to reach a consensus on the map. Figure \ref{fig:hallway_26_0} and \ref{fig:hallway_26_1} show the final belief of the hallways configuration for the example robot for plain consensus and W-MSR consensus respectively. It is clear that the non-cooperative robot is able to prevent exploration of the top section of the hallway when the robots are using plain consensus, but is unable to do so when the robots are running the resilient W-MSR. The cost for resilience is that W-MSR takes significantly longer to converge since new information takes longer to propagate through the graph and because it requires more than one robot to measure each cell in the environment. 

\section{Resilient Consensus for Enhanced Voronoi Graphs}
\label{sec:enhanced}
Numerical studies were conducted to explore the $(r,s)$-robustness gained by continuing to connect more distant neighbors in $\calG_\Delta$. To this end we define $\calG_{\Delta K}$ as the graph created from $\calG_\Delta$ by connecting $k$-hop neighbors for $k=2,...,K$, or equivalently connecting the 2-hop neighbors of $\calG_{\Delta K-1}$. In the first numerical study, randomly generated graphs were created by choosing $N$ robots, $N\in\brl{10,19}$, from a uniform distribution, and distributing them uniformly at random locations within a rectangular environment which has been randomly scaled to produce rectangles ranging from almost square to long and thin. From the generated robot positions, the corresponding Voronoi graph $\calG_\Delta$ was constructed. Following Definition \ref{defn:rsrobust} and beginning with $r=s=\lceil N/2 \rceil$, every pair of disjoint subsets of vertices $\crl{\calS_1,\calS_2}$ was checked to verify $(r,s)$-robustness of the graph. In the case where a pair of sets did not meet the conditions in Definition \ref{defn:rsrobust} for the current $r=s$, $r$ and $s$ were decremented until the conditions were met. This resulted in an algorithm that found the maximum $(r,s)$-robustness such that $r=s$ for each graph. Since $(F+1,F+1)$-robustness can reject $F$-global malicious robots, the largest $r=s$ is sufficient to understand to what extent the graph is robust to this threat model. Graphs with $N\geq 20$ were not tested due to the exponentially increasing run time as the number of pairs of disjoint subsets of vertices scales with $\calO(3^N)$. Table \ref{table:sim_rs_extention} shows the results of 100 random graphs generated for each graph type. 100 graphs were generated for each of the graph types ($\calG_\Delta$, $\calG_{\Delta 2}$, etc.). Table \ref{table:sim_rs_extention} shows the percentage of each graph type which achieved $(r,s)$-robustness for each $r=s$. Note that graphs which are $(r,s)$-robust are also $(r-l,s-l)$-robust.

\begin{table}[h!]
\centering
\setlength{\tabcolsep}{3pt}
\begin{tabular}{r | c c c c c c c c c | c} 
 & \multicolumn{9}{|c|}{$r=s$} & \\
   & 2 & 3 & 4 & 5 & 6 & 7 & 8 & 9 & 10 & KN \\ 
 \hline
 $\calG_\Delta$  & 100\% &      &      &      &      &      & & & & 11\% \\ 
 $\calG_{\Delta 2}$       &       & 100\% & 81\% & 18\% & 1\%  &      & & & & 26\% \\
 $\calG_{\Delta 3}$    &       &      & 100\% & 82\% & 64\% & 38\% & 14\% & 2\% & & 69\% \\
 $\calG_{\Delta 4}$    &       &      &      & 100\% & 75\% & 62\% & 47\% & 33\% & 12\% & 99\% \\
 \hline
\end{tabular}
\caption{The percentage of randomly generated graphs which have particular $(r,s)$-robustness assuming $r=s$. KN indicates the percentage of the randomly generated graphs that are complete, i.e., have all possible edges.}
\label{table:sim_rs_extention}
\end{table}

It can be seen that as expected from the theorems and proofs provided in Section~\ref{sec:resilient_voronoi_theorems}, the $\calG_\Delta$ graphs are always $(2,2)$-robust and the $\calG_{\Delta 2}$ graphs are always at least $(3,3)$-robust. Interestingly, the $(r,s)$-robustness of $\calG_{\Delta 2}$ graphs is frequently greater than $(3,3)$-robust. It can also be seen that as more distant neighbors are connected, the minimum $(r,s)$-robustness increases linearly. The final column shows the percentage of the generated graphs which are complete graphs. The number of robots simulated is limited due to the computational complexity of checking for $(r,s)$-robustness\cite{Zhang2012_complexity_of_rrobust}, so the random examples quickly become uninteresting as the connectivity increases since by $\calG_{\Delta 4}$ most of the generated graphs are complete.

To explore the limits of the graph extension method, it is more helpful to look at a low connectivity case. The ``two lines'' example in Figure \ref{fig:special_voronoi_formations_twolines} provides just such an case. In this example, $\calG_\Delta$ is the same graph that would be constructed if starting with a single triangle face on three vertices, each additional vertex is added to maximize the average shortest path from the current vertices to the new vertex. In this case, we find a linear relationship between $(r,s)$-robustness and the level of hops which are connected from the base graph $\calG_\Delta$, with the single caveat that there must be sufficient vertices, e.g., a five-vertex graph cannot be $(4,4)$-robust no matter how many hops away vertices are connected, this requires at least seven vertices. 

Table \ref{table:sim_twolines} shows how the $(r,s)$-robustness increases as more distant neighbors in $\calG_\Delta$ are connected. The formation was tested with $N=11$ (the smallest number for $(6,6)$-robust) and $N=19$ to show how the number of edges scales as more vertices are added. The $(r,s)$-robustness an be seen to increase linearly with $K$ for $\calG_{\Delta K}$, providing further evidence for the proposed method of increasing robustness. Also of note is that for the two lines formation the number of edges for a given $K$ scales roughly linearly with the number of robots. Other formations will have different scaling properties, but this indicates that at least for some formations $\calG_{\Delta K}$ provides a communication graph that scales well with the number of robots and the desired $(r,s)$-robustness.

\begin{table}[h!]
\centering
\setlength{\tabcolsep}{3pt}
\begin{tabular}{r | c c c c c c c c} 
  & $\calG_\Delta$ & $\calG_{\Delta 2}$ & $\calG_{\Delta 3}$ & $\calG_{\Delta 4}$ & $\calG_{\Delta 5}$ & $\calG_{\Delta 6}$ & $\calG_{\Delta 7}$ & $\calG_{\Delta 8}$\\ 
  \hline
$N=11$: $\;\;r=s$ & 2 & 3 & 4 & 5 & 6 & 6 & 6 & 6\\
\# edges & 19 & 34 & 45 & 52  & 55  & 55 & 55 & 55\\
\hline
$N=19$: $\;\;r=s$ & 2 & 3 & 4 & 5 & 6 & 7 & 8 & 9\\
\# edges & 35 & 66 & 93 & 116 & 135 & 150 & 161 & 168\\
\end{tabular}
\caption{$(r,s)$-robustness for the ``two lines'' formation on 10 vertices and 19 vertices. For context, the complete graph on 11 vertices, K11, has 55 edges and K19 has 171. $r=s=6$ for $N=11$ in all graphs after $\calG_{\Delta 5}$ since at that point the graph is always complete and so with no possible edges to add, its robustness cannot be further increased.}
\label{table:sim_twolines}
\end{table}

\section{Conclusions}
In this paper, we prove that the communication graph built by connecting Voronoi neighbors is resilient to a single F-global non-cooperative team member. We further prove that this resilience can be improved by connecting 2-hop neighbors in order to reject 2 non-cooperative robots in the network or 1 non-cooperative robot in the neighborhood of every cooperative robot. We also provide numerical evidence that this resilience can be further improved by continuing to connect more distant neighbors in the Voronoi communication graph. This work opens the door to the use of resilient consensus with many coverage and exploration algorithms that already assume the use of Voronoi tessellations and/or Voronoi communication networks and also provides a new method for constructing resilient networks for other applications.  We show several examples of the use of Voronoi communication graphs and their extended versions in domains where direct verification of robustness would be prohibitively expensive such as in large stationary networks, and continuously varying networks. Finally, we show a complex example of consensus on an occupancy grid map, where resilient consensus allows the robots to fully explore the environment despite a malicious team member.

The results in this paper suggest several directions for future research. One interesting direction is the application of resilient consensus to tasks such as distributed filtering and distributed target tracking. Tasks for which information is privileged based on sensing location, such as in target tracking, will require new ways of redundantly partitioning work to ensure that enough information is available to run resilient consensus. A resilient graph will not be enough if only a single robot has information to share. This challenge also appears in the mapping example in this paper, where the naive solution of allowing each robot to seek information that it doesn't have results in significant clustering as large groups of robots move together to explore the remaining uncertain cells. Resilient distributed filtering, such a resilient form of the distributed Kalman filter~\cite{Olfati-Saber2007_kalman}, also relies on consensus, so future research might examine the effect of using W-MSR consensus with such algorithms. Finally, since this work relies on computing Voronoi neighbors, it would be useful to study the interaction between the results presented here and existing distributed Voronoi partition algorithms and the distributed construction of resilient Voronoi communication graphs.

\appendix[Proof: \texorpdfstring{$(3,3)$}{(3,3)}-robustness of \texorpdfstring{$\calG_{\Delta 2}$}{extended graph}]
\begin{IEEEproof}
First we will cover some notation that is used throughout the proof. First, it is often necessary to indicate edges and neighbors that are in the subgraph $\calG_\Delta$, the Delaunay triangulation that was used to create $\calG_{\Delta 2}$. We refer to these edges as $\Delta$-edges, and neighbors as $\Delta$-neighbors, with a $\Delta$-path being a path using only $\Delta$-edges. Further, we use the notation $\calN_{j,\Delta}$ to indicate the set of $v_j$'s $\Delta$-neighbors and $\calN_j$ to refer to the set of all $v_j$'s neighbors. 

Second, we will frequently make use of the property that the $\Delta$-neighbors of a vertex $v_j$ form a simple path in $\calG_\Delta$. This is also used in the proof for Theorem \ref{thm:voronoi_robust}. We also occasionally use the definition of triangulation as a graph with all triangular faces except possibly the outer face which is a cycle. 

For this proof we will break the cases for $\calS_1$ and $\calS_2$ into two main cases. The first case is when the connected components of either $\calG_{\Delta 2}[\calS_1]$ or $\calG_{\Delta 2}[\calS_2]$ contain no edges from $\calE_\Delta$, or consist of exactly two vertices with a connection by an edge in $\calE_\Delta$. This case is handled separately because it encapsulates all cases where Conditions 1 and 2 from Definition \ref{defn:rsrobust} are required. The second case is where for both $\calG_{\Delta 2}[\calS_1]$ and $\calG_{\Delta 2}[\calS_2]$ there is at least one connected component which contains a $\Delta$-path of more than 3 vertices, or a connected component with two $\Delta$-connected vertices  and at least 1 in-set extended edge. In this second case, we will instead show that Condition 3 of Definition \ref{defn:rsrobust} holds.

\begin{enumerate}
    \item The connected components of either $\calG_{\Delta 2}[\calS_1]$ or $\calG_{\Delta 2}[\calS_2]$ contain no edges from $\calE_\Delta$, or consist of exactly two vertices with a connection by an edge in $\calE_\Delta$. Call the set for which this is true $\calS_a\in\crl{\calS_1,\calS_2}$. In this case we will show that every vertex in $\calS_a$ is in $\calX_{\calS_a}^3$ which then satisfies Condition 1 or 2 from Definition \ref{defn:rsrobust}.
    \begin{figure}
        \centering
        \begin{subfigure}[t]{.10\textwidth}
        \centering
        \includegraphics[width=\textwidth]{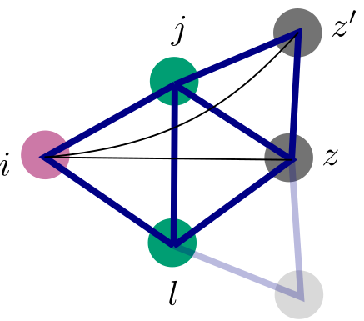}
        \caption{$v_i\in\calS_{a,E}$}
        \label{fig:proof33_case1a}
        \end{subfigure}
        \begin{subfigure}[t]{.20\textwidth}
        \centering
        \includegraphics[width=\textwidth]{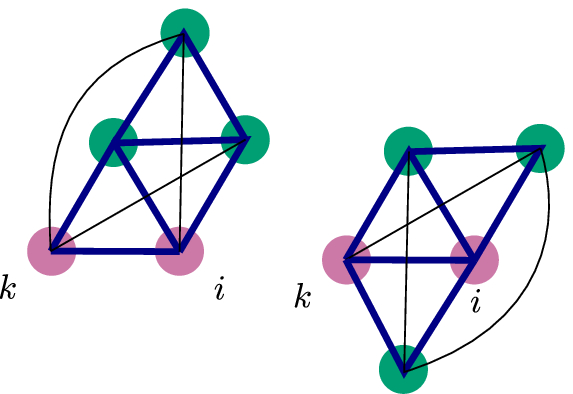}
        \caption{$v_i,v_k$ both have no other in-set neighbors}
        \label{fig:proof33_case1b}
        \end{subfigure}
        \caption{For case 1: Pink vertices indicate $\calS_1$, green $\bar{\calS_1}$, and grey could be either. Thick blue edges are $\Delta$-edges, whereas thin black edges are extended edges.}
        \label{fig:proof33_case1}
    \end{figure}
    \begin{enumerate}
        \item Consider any connected component of $\calG_{\Delta 2}[\calS_a]$ which contains no $\Delta$-edges. Denote the set of vertices in this connected component as $\calS_{a,E}\subset\calS_a$.
        Consider vertex $v_i\in\calS_{a,E}$. Since $v_i\in\calS_{a,E}$, we know it has no $\Delta$-edges to any other vertices in $\calS_a$. Since $\calG_\Delta$ is formed via Delaunay triangulation, we know all the faces are triangles. Thus $v_i$ is the vertex of at least one triangle and the other two vertices of this triangle, denoted $v_j, v_l$, must be in $\bar{\calS_a}$, the complement of $\calS_a$. If $v_i$ is the vertex of more than one triangle, then $v_i\in\calX_{\calS_a}^3$. If $v_i$ is the vertex of only a single triangle face, then since we also know that $|\calV_\Delta|\geq 5$, the graph contains at least three triangle faces. By our assumption, none of the remaining triangles can include $v_i$. $v_j, v_l$ must be part of a second triangle with a shared neighbor $v_z$. The third triangle could share edge $(v_j,v_z)$, or $(v_l,v_z)$. Note that adding a triangle face which includes the edge $(v_j,v_l)$ again does not result in a Delaunay triangulation because a single edge cannot be shared by more than two triangle faces. Denote the third vertex of this triangle as $v_z'$. We know $v_z',v_z\in\calN_{i}$ via extended, 2-hop edges since each shares at least one $\Delta$-neighbor with $v_i$. We know that $v_z,v_z'$ cannot both be in $\calS_{a,E}$ since $v_z'\in\calN_{z,\Delta}$ (and vise versa) and $\calS_{a,E}$ is defined as having no in-set $\Delta$-edges. Therefore at most one $v_z$ or $v_z'$ is in $\calS_a$. That leaves at least three nodes $v_j,v_l,v_z$ (or $v_z'$)$\in\calN_{i}\cap\bar{\calS_a}$, so $v_i\in\calX_{\calS_a}^3$. Thus $\forall i\in\calS_{a,E}$ and $\forall\calS_{a,E}\in\calS_a$, $i\in\calX_{\calS_a}^3$.
        \item Consider any connected component of $\calG_{\Delta 2}[\calS_a]$ which contains exactly two vertices connected by a $\Delta$-edge. Denote these vertices $v_i,v_k$. We know that the edge $(v_i,v_k)$ is part of at least one and at most two triangle faces. We also know that there are at least five vertices in the graph, and therefore at least three triangles. No matter how these three triangles are arranged, the resulting extended graph on those vertices is complete, and thus within those 5 vertices no other extra vertices in $\calS_a$ are allowed, as their existence would violate the assumption that $v_i,v_k$ have no other in-set neighbors. Therefore, since only $v_i,v_k\in\calS_a$ in that set of five fully connected vertices, $|\calN_i\cap\bar{\calS_a}|\geq 3$ and  $|\calN_k\cap\bar{\calS_a}|\geq 3$, implying $v_i,v_k\in\calX_{\calS_a}^3$, $\forall v_i,v_k$ connected by a $\Delta$-edge with no other in-set edges.
        To sum up, if $\exists \calS_a\in\crl{\calS_1,\calS_2}$ such that $\calG_{\Delta 2}[\calS_a]$ contains only connected components that either have no $\Delta$ edges, or are exactly two vertices connected with a $\Delta$ edge, then all vertices in that set are in $\calX_{\calS_a}^3$. Therefore in this case Condition 1 (for $\calS_a=\calS_1$) or Condition 2 (for $\calS_a=\calS_2$) is satisfied for $r=3$.
    \end{enumerate} 
    \item The remaining case is that in both $\calG_{\Delta 2}[\calS_1]$ and $\calG_{\Delta 2}[\calS_2]$ there is at least one connected component which contains a $\Delta$-path of more than three vertices, or a connected component with two $\Delta$-connected vertices  and at least one extended edge. We will show that in this case $\calS_1$ has at least two vertices in $\calX_{\calS_1}^3$ and since we have the same assumption made about both sets we can then apply the same logic to $\calS_2$ which allows us to conclude that $|\calX_{\calS_1}^3|+|\calX_{\calS_2}^3|\geq 4$. Since $4\geq 3$, this more than satisfies Condition 3 for $r=3$ and $s=3$.
    
    Consider a vertex $v_s\in\calS_1$ such that $|\calS_1\cap\calN_{s,\Delta}|\neq \emptyset$ and a vertex $g\in\calS_2$ which belongs to a connected component of $\calG_\Delta[\calS_2]$ with 3 or more vertices, or of only 2 vertices at least one of which has an in-set extended edge. We know there is a $\Delta$-path between these two vertices since $\calG_\Delta$ is connected. Since this path starts in $\calS_1$ and ends in $\calS_2$, we know there must be a pair of vertices on the path, $v_i,v_j$, for which $v_i\in\calS_1$ and $v_j\in\bar{\calS_1}$.
    Thus there is a $\Delta$-path $\prl{v_s,...,v_k, v_i, v_j,...,v_g}$ for which $v_s,v_k,v_i\in\calS_1$ (with possibly $v_s=v_k$) belong to a connected component of $\calG_\Delta[\calS_1]$ with at least two vertices. This path can also always be defined such that $\prl{v_j,...,v_g}\in\bar{\calS_1}$ as shown in Figure \ref{fig:proof33_path}. To see this, consider $\prl{v_j,...,v_g}\notin\bar{\calS_1}$ so there $\exists v_m\in\calS_1$ on the path between $v_j$ and $v_g$. If $\calN_{m,\Delta}\cap\calS_1=\emptyset$, then we know all $\calN_{m\Delta}\in\bar{\calS_1}$. Since the $\Delta$-neighbors of every vertex in $\calG_\Delta$ form a $\Delta$-path, we then know that there is a $\Delta$-path in $\bar{\calS_1}$ which can circumvent $v_m$ to create a path $\prl{v_j,...,\prl{\subset\calN_{m\Delta}},...,v_g}\in\bar{\calS_1}$. Alternatively if $|\calN_{m,\Delta}\cap\calS_1|\geq 1$, then $v_m$ is part of a connected component of $\calG_\Delta[\calS_1]$ with 2 or more vertices and we can redefine $v_s=v_m$ and obtain a path where $v_m$ is not between $v_j$ and $v_g$. These cases are shown in Figure \ref{fig:proof33_pathm}. Therefore there exists a $\Delta$-path $\prl{v_s,...,v_k, v_i, v_j,...,v_g}$ for which $\prl{v_s,...,v_k,v_i}\in\calS_1$,  $\prl{v_j,...,v_g}\in\bar{\calS_1}$, and due to the definitions of the connected components $v_s$ and $v_g$ are chosen from, $|\prl{v_s,...,v_k,v_i}|\geq 2$ and $|\prl{v_j,...,v_g}|\geq 3$ or $|\prl{v_j,...,v_g}| = 2$ with an extended edge from $v_g$ to another vertex in $\calS_2$. Note: The purpose of this path is to provide a reasonable location for finding two vertices in $\calX_{\calS_1}^3$, since near the transition $v_i,v_j$ there are at least 2 vertices in $\calS_1$ and at least 3 nearby vertices in $\bar{\calS_1}$. There are two cases to consider.
    
    \begin{figure}
        \centering
        \begin{subfigure}[b]{0.5\textwidth}
        \centering
            \includegraphics[width=0.7\textwidth]{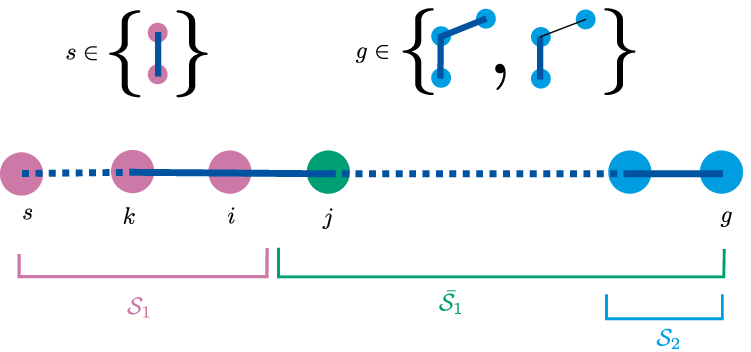}
            \caption{}
            \label{fig:proof33_path}
        \end{subfigure}
        \par\bigskip
        \begin{subfigure}[b]{0.5\textwidth}
        \centering
            \includegraphics[width=0.7\textwidth]{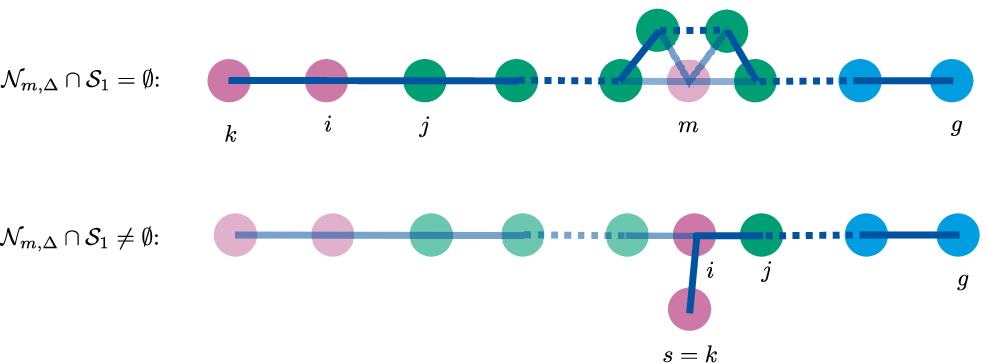}
            \caption{}
            \label{fig:proof33_pathm}
        \end{subfigure}
        \caption{(a) The path $\prl{v_s,...,v_k,v_i,v_j,...,v_g}$ (b) A vertex $v_m\in\calS_1$ between $v_j$ and $v_g$ on the path does not prevent the path from being defined as having only vertices in $\calS_1$ before $v_j$ and only vertices in $\bar{\calS_1}$ after. Here pink vertices are in $\calS_1$, green in $\bar{\calS_1}$, and blue in $\calS_2$. Thick blue edges are $\Delta$-edges, while thin black edges are extended edges added between 2-hop neighbors. Dotted edges indicate a $\Delta$-path. }
        \label{fig:proof33_pathdetails}
    \end{figure}
    \begin{enumerate}
        \item There are at least three vertices in the path $\prl{v_j,...,v_g}$. This is the case when $v_g$ belongs to a connected component of $\calG_\Delta[\calS_2]$ of size three or more, or when $v_g$ belongs to a connected component of only two vertices (with a required extended edge), but some vertices in $\prl{j,...g}\notin\calS_2$, i.e., there are some vertices in that are in neither set along the $\Delta$-path. In this case, we will denote the 2nd and 3rd vertices on the path $l,q\in\bar{\calS_1}$, and consider the sub-path $\prl{k,i,j,l,q}$. There are three cases to consider.
        \begin{enumerate}
            \item $|\calN_{i,\Delta}\cap\bar{\calS_1}|\geq 3$: In this case $v_i$ has at least 3 $\Delta$-neighbors in $\bar{\calS_1}$ so $v_i\in\calX_{\calS_1}^3$. Also, $v_k$ has at least extended edges to each of those 3 neighbors, so $v_k\in\calX_{\calS_1}^3$. Therefore $|\calX_{\calS_1}^3|\geq 2$.
            \item $|\calN_{i,\Delta}\cap\bar{\calS_1}|=2$: One of these neighbors is $v_j$, lets call the other $v_j'$, with $v_j'\neq v_j$. Right away we can see that $v_i\in\calX_{\calS_1}^3$, with neighbors $\crl{v_j,v_j',v_l}$ (or if $v_j'=v_l$, $\crl{v_j,v_l,v_q}$). We know $\calN_{l,\Delta}$ form a $\Delta$-path, so $v_j$ must share a $\Delta$ neighbor $v_b$ with $v_l$. The following cases are shown in Figure \ref{fig:proof33_3row_ii}.
            \begin{enumerate}
                \item if the only option is $v_b=v_i$ (i.e., $\calN_{j,\Delta}\cap\calN_{l,\Delta}=\crl{i}$, shown in Figure \ref{fig:proof33_3row_ii_a}) then there is a path in $v_l$'s $\Delta$-neighbors from $v_i$ to $v_q$ that does not include $v_j$ (To see this note that $v_j$ has only a single $\Delta$-neighbor in $\calN_{l,\Delta}$ and so must form a path endpoint). There are transition vertices in this path $v_m\in\calS_1\cap\calN_{l,\Delta}$ and $v_n\in\bar{\calS_1}\cap\calN_{l,\Delta}\cap\calN_{m,\Delta}$. $v_m\in\calX_{\calS_1}^3$ with neighbors $\crl{v_j,v_l,v_q}$. $v_m\neq v_i$ (since $v_m=v_i$ would have $|\calN_{i,\Delta}\cap\bar{\calS_1}|\geq 3$, the previous case) so $|\calX_{\calS_1}^3|\geq 2$.
                \item $v_b\neq v_i$, $v_b\in\calS_1$ (Figure \ref{fig:proof33_3row_ii_b}), $v_b\in\calX_{\calS_1}^3$ with neighbors $\crl{v_j,v_l,v_q}$ so including $v_i$, one has $|\calX_{\calS_1}^3|\geq 2$.
                \item $v_b\neq v_i$, $v_b\in\bar{\calS_1}$. First consider the case that $v_j$ has a neighbor in $\calS_1\setminus\crl{i}$ (Figure \ref{fig:proof33_3row_ii_c1}), then that neighbor and $v_i$ are in $\calX_{\calS_1}^3$. If we assume that neighbor doesn't exist, consider that $\calN_{i,\Delta}$ also must form a path (Figure \ref{fig:proof33_3row_ii_c2}). Therefore there must be a $\Delta$-path in $\calN_{i,\Delta}$ from $v_k$ to $v_j$ but since we have assumed that $v_j$ as no $\Delta$-neighbors in $\calS_1$, and $v_i$ only has two $\Delta$-neighbors in $\bar{\calS_1}$, $v_j'$ is the only possible $\Delta$-neighbor for $j$ in $\calN_{i,\Delta}$ so $j\in\calN_{j',\Delta}$, and the path of $v_i$'s neighbors must go from $v_k$ to $v_j'$ without passing $v_j$. This path contains $v_k\in\calS_1$, and $v_j'\in\bar{\calS_1}$ there must be a path in $\calN_{i,\Delta}$ from $v_k$ to $v_j'$ with transition nodes $v_m'\in\calS_1\cap\calN_{i,\Delta}$ and $v_n'\in\bar{\calS_1}\cap\calN_{i,\Delta}\cap\calN_{m',\Delta}$. We know $v_n'=v_j'$ since $v_i$ can only have the two $\Delta$-neighbors in $\bar{\calS_1}$. Furthermore, consider that $\calN_{j,\Delta}$ must also from a $\Delta$-path and since $v_i$ is $v_j$'s only $\Delta$-neighbor in $\calS_1$, all nodes in $\calN_{j,\Delta}\setminus\crl{v_i}\in\bar{\calS_1}$. $v_i$ cannot have any more neighbors in $\bar{\calS_1}$, so there must be a path in $\bar{\calS_1}\cap\calN_{j,\Delta}$ from $v_j'$ to $v_b$. We can then see that $v_m'\in\calX_{\calS_1^3}$ with neighbors $v_j',v_j$ plus a node on this path to $v_b$. Therefore in this case $|\calX_{\calS_1}^3|\geq 2$ with nodes $v_m'$ and $v_i$.
            \end{enumerate}
            \item $|\calN_{i,\Delta}\cap\bar{\calS_1}|=1$: (Figure \ref{fig:proof33_3row_iii}) We will show this case is included in the previous cases via a relabeling of vertices. Though in this case, $v_i$ has only a single neighbor in $\bar{\calS_1}$, $\calN_{j,\Delta}$ must form a $\Delta$-path so there must be a path in $\calN_{j,\Delta}$ between $v_i$ and $v_l$. Since $v_i\in\calS_1$, $v_l\in\bar{\calS_1}$ there must be at least one transition between sets $v_m\in\calS_1\cap\calN_{j,\Delta}$ and $v_n\in\bar{\calS_1}\cap\calN_{j,\Delta}\cap\calN_{m,\Delta}$. Choose $v_m,v_n$ to indicate the transition such that $\prl{i,...,m}\in\calS_1$, and consider this new path $\prl{s,...,k,i,...,m,n,j,l,q,...,g}$. A relabeling of $v_i=v_m$ and the vertex proceeding $v_m$ in $\prl{i,...,m}$ as $v_k$, there is a path $\crl{s,...,k,i,j,l,q,...,g}$ where $|\calN_{i,\Delta}\cap\bar{\calS_1}|\geq 2$ which using the previous two cases implies that $|\calX_{\calS_1}^3|\geq 2$.
        \end{enumerate}
        \begin{figure}[ht]
            \centering
            \parbox{0.4\textwidth}{
                \parbox{.15\textwidth}{%
                \centering
                    \subcaptionbox{$b=i$.\label{fig:proof33_3row_ii_a}}{\includegraphics[width=\hsize]{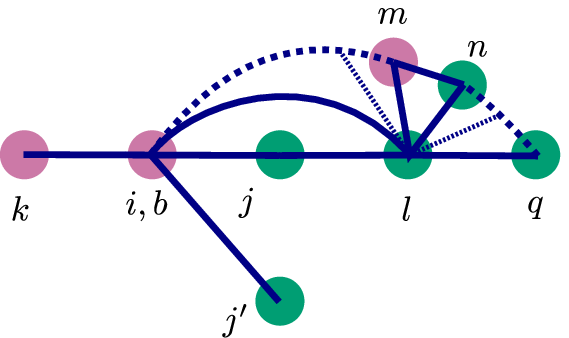}}
                    \subcaptionbox{$b\neq i$ and $b\in\calS_1$.\label{fig:proof33_3row_ii_b}}{\includegraphics[width=\hsize]{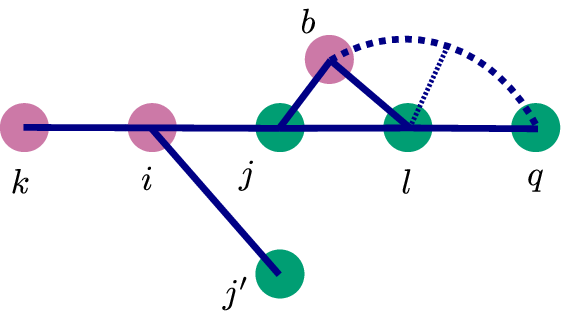}}
                    \vskip1em
                    \subcaptionbox{$b\neq i$ and $b\notin\calS_1$ with $|\calN_{j,\Delta}\cap\calS_1|\geq 2$.\label{fig:proof33_3row_ii_c1}}{\includegraphics[width=\hsize]{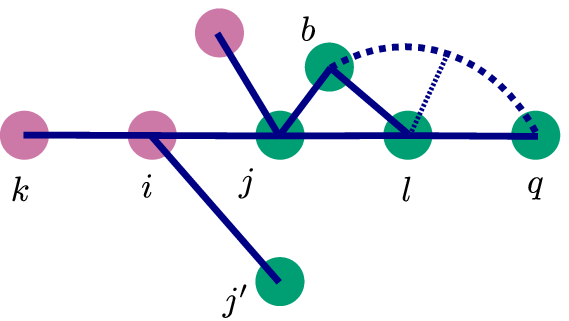}}  
                }
                \hskip1em
                \parbox{.20\textwidth}{%
                \centering
                    \subcaptionbox{$b\neq i$ and $b\notin\calS_1$ with $\calN_{j,\Delta}\cap\calS_1=\crl{i}$.\label{fig:proof33_3row_ii_c2}}{\includegraphics[width=\hsize]{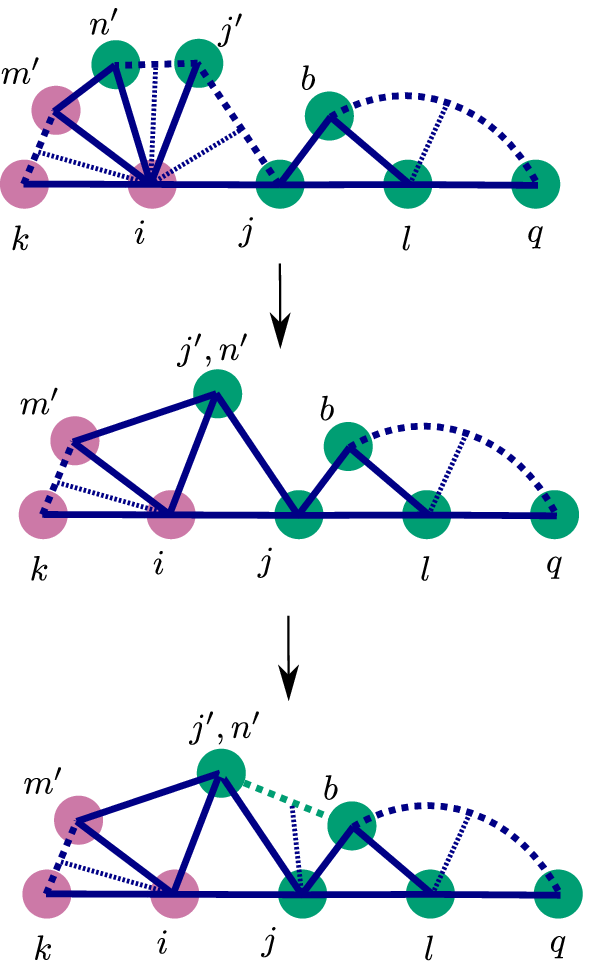}}
                }
            }
            \caption{Cases for $ii)$ where $i$ has two neighbors.}
            \label{fig:proof33_3row_ii}
        \end{figure}
        \begin{figure}
            \centering
            \includegraphics[width=0.2\textwidth]{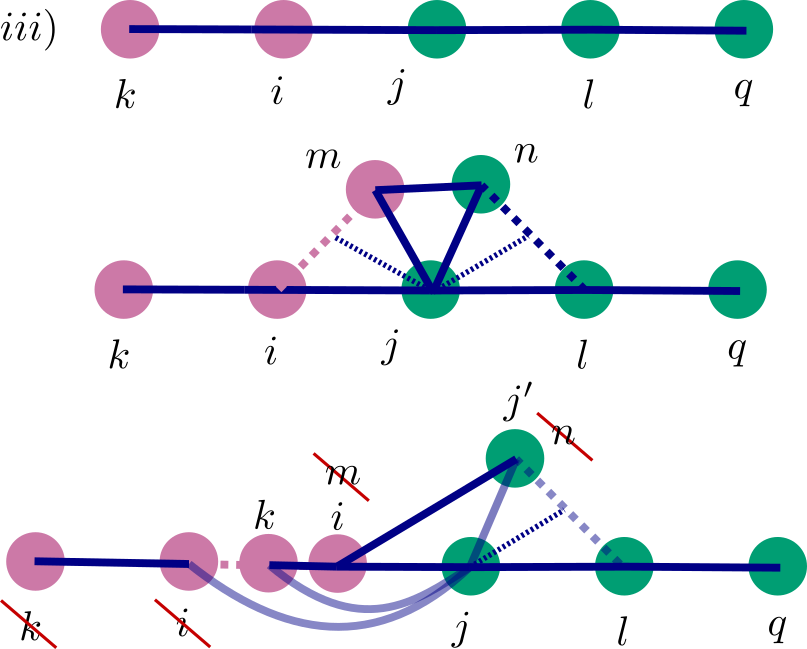}
            \caption{The case where $v_i$ only has one $\Delta$-neighbor in $\bar{\calS_1}$ is included in the cases where $v_i$ has two or more $\Delta$-neighbors in $\bar{\calS_1}$ via relabeling of nodes.}
            \label{fig:proof33_3row_iii}
        \end{figure}
        \item There are not three vertices in the path $\prl{v_j,...,v_g}$. In this case there must be two vertices such that $v_j\in\calS_2$ and $v_g\in\calS_2$ are a pair of $\Delta$-connected vertices which have a neighbor in $\calS_2$ via an extended edge. In this case we call the vertex connected with the extended edge as $v_r\in\calN_{l}\cap\calS_2$ and assume without loss of generality that it is connected to $v_g$ (there is a path in $\calS_1\cap\calN_{j,\Delta}$ from $v_i$ to $v_g$, so if the extended edge is connected to $v_j$, the case is identical after a relabeling of vertices). Since $v_r$ and $v_g$ are connected with an extended edge, we know by the construction of $\calG_{\Delta 2}$ that they must share a $\Delta$-neighbor, $v_m$. If $v_m\in\bar{\calS_1}$, this case is covered by the previous case with $v_q=v_m$. So assume $v_m\in\calS_1$. $v_m\in\calX_{\calS_1}^3$ with neighbors $\crl{j,g,r}$. Figure \ref{fig:proof33_2extrow} is for visualizing these steps.  $\calN_{m,\Delta}$ must form a $\Delta$-path and $v_g$ does not connect via a $\Delta$-connection to $v_r$, so there must be at least one intermediate node $v_a\in\calN_{g,\Delta}\cap\calN_{m,\Delta}$. If $v_a\in\bar{\calS_1}$, then there are three vertices in $\bar{\calS_1}$ in a row, and that is covered in the previous case. Otherwise, if $v_a\in\calS_1$, $v_a\in\calX_{\calS_1}^3$ with neighbors $\crl{v_j,v_g,v_r}$, thus $|\calX_{\calS_1}^3|\geq 2$ with nodes $v_a,v_m$.
        \begin{figure}
            \centering
            \includegraphics[width=0.4\textwidth]{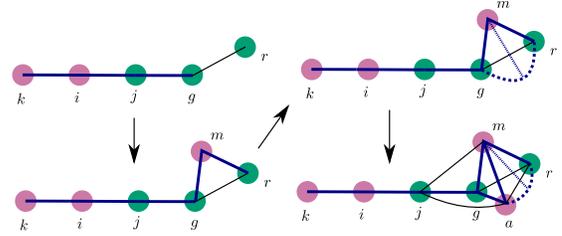}
            \caption{The case where there are only 2 vertices in the path $\prl{v_j,...,v_g}$ and an extended edge to $v_r\in\calS_2$.}
            \label{fig:proof33_2extrow}
        \end{figure}
    \end{enumerate}
    
    We conclude that $|\calX_{\calS_1}^3|\geq 2$ in the case where both $\calG_\Delta[\calS_1]$ and $\calG_\Delta[\calS_2]$ have at least one connected component with 3 or more vertices, or a connected component with 2 vertices at least one of which has an in-set extended edge in $\calG_{\Delta 2}$. Using the same logic we can conclude the same for $\calX_{\calS_2}^3$. Thus we have shown that $|\calX_{\calS_1}^3|+|\calX_{\calS_2}^3| \geq 4\geq 3$ and so Condition 3 is satisfied for $r=3$, $s=3$.
\end{enumerate}
Finally, since we have shown that for all possible choices of disjoint subsets $\calS_1$ and $\calS_2$ one of the three Conditions of Definition \ref{defn:rsrobust} holds for $r=3$ and $s=3$, we conclude that the extended graph $\calG_{\Delta 2}$ is $(3,3)$-robust.
\end{IEEEproof}

\addtolength{\textheight}{-12cm}   






\bibliographystyle{cls/IEEEtran}
\footnotesize{
\bibliography{bib/IEEEabrv,bib/library}
}



\end{document}